\documentclass{article}

\usepackage[utf8]{inputenc}
\usepackage{graphicx}
\usepackage{multicol}
\usepackage{subcaption}
\usepackage[linesnumbered,ruled,vlined]{algorithm2e}
\usepackage{multirow}
\usepackage{url}

\begin{document}
\title{Explainable Data Imputation using Constraints}
\pagestyle{plain}
 \author{Sandeep Hans \\{\small shans001@in.ibm.com}  \\{\small IBM Research} 
   \and Diptikalyan Saha  \\{\small diptsaha@in.ibm.com}   \\{\small IBM Research} 
   \and Aniya Aggarwal  \\{\small aniyaagg@in.ibm.com}  \\{\small IBM Research} 
}%
 \date{}
\maketitle

\begin{abstract}
    Data values in a dataset can be missing or anomalous due to mishandling or human error. Analysing data with missing values can create bias and affect the inferences. Several analysis methods, such as principle components analysis or singular value decomposition, require complete data. Many approaches impute numeric data and some do not consider dependency of attributes on other attributes, while some require human intervention and domain knowledge. We present a new algorithm for data  imputation based on different data type values and their association constraints in data, which are not handled currently by any system. We show experimental results using different metrics comparing our algorithm with state of the art imputation techniques. Our algorithm not only imputes the missing values but also generates human readable explanations describing the significance of attributes used for every imputation.
\end{abstract}
\maketitle

\newcommand{\cnn}{CAT-NUM-NUM }
\newcommand{\acatnum}{CAT-NUM }
\newcommand{\catnum}{CAT\_NUM }
\newcommand{\catcat}{CAT-CAT }
\newcommand{\numnum}{NUM-NUM }
\newcommand{\ctext}{TEXT }
\newcommand{\num}{NUMERIC }
\newcommand{\float}{FLOAT }
\newcommand{\cattext}{CAT\_TEXT }
\newcommand{\acattext}{CAT-TEXT }
\newcommand{\cempty}{EMPTY }
\newcommand{\cdate}{DATE }
\newcommand{\datedate}{DATE-DATE }
\section{Introduction}

Many real-world datasets may contain missing values for various reasons. 
Training a model with a dataset that has a lot of missing values can drastically impact the machine learning model’s quality. Some algorithms assume that all values are available and hold meaningful value.
One way to handle this problem is to get rid of all the observations having any value missing. However, it involves the risk of losing data points with valuable information. The best strategy is to impute these missing values. 
%
However, most of the imputation techniques impute values for an attribute that may not conform with other attributes.
 For example, while imputing salary for an employee, the technique may not consider the designation of the employee and thus, impute a non-conforming value w.r.t designation attribute although the value is valid .
 
Some naive solutions for this problem are removing the rows containing missing values, substituting the missing values with mean or median of non-missing values of the attribute, or most frequent value for categorical data. There are certain problems with these techniques. For example, if you remove rows containing missing values  other algorithms will throw errors complaining about the missing values. In that case, you will need to handle the missing data and clean it before feeding it to the algorithm. Rows of a data are often not complete, especially when dealing with heterogeneous data sources. Discarding an entire row of a table if just one column has a missing value would often discard a substantial part of the data. Substituting the missing value of a numerical attribute by mean/median of non-missing values of the attribute doesn’t factor the correlations between features. It only works on the column level and gives poor results on encoded categorical features. It is also not very accurate, can conflict with other attributes and doesn’t account for the uncertainty in the imputations. Substituting the missing value of a categorical attribute by most frequent value of the attribute also doesn’t factor the correlations between features and can introduce bias in the data.

 In this paper, we focus on given a dataset with missing values, substitute the missing data with the values which conforms with the rest of the data. Our technique inherently provides explanations for each imputation done, which can be used further to explain the results of the task at hand like drop in accuracy of an AI model or increase in bias in the data.

We present an algorithm for data  imputation based on different data type values and their association constraints in data, which are not handled currently by any system. We also show experimental results comparing our algorithm with state of the art imputation techniques. Our algorithm not only imputes the missing values but also generates human readable explanations for each imputation,  describing the significance of other attributes used for the imputation.

\section{Related Work}

Most research in the field of imputation focuses on imputing missing values in matrices, that is imputation of numerical values from other numerical values. Popular approaches include k-nearest neighbors (KNN)~\cite{knn-aai03}, multivariate imputation by chained equations (MICE)~\cite{little2002statistical}, matrix factorization~\cite{KorenBV09,MazumderHT10,TroyanskayaCSBHTBA01} or deep learning methods~\cite{datawig-cikm18,datawig-jmlr19,mida-pakdd18,zhang-arxiv18,miwai-icml19} . While some recent work addresses imputation for more heterogeneous data types~\cite{missforest-bioinformatics12,gain-icml18,vae-arxiv18}, heterogeneous in those studies refers to binary, ordinal or categorical variables, which can be easily transformed into numerical representations. 

K-nearest neighbors (KNN) based data imputation~\cite{knn-aai03} replaces the missing data for a given variable by averaging (non-missing) values of its neighbors. Fuzzy K-means based data imputation: an extension of KNN based on fuzzy K-means clustering
This works for numerical data only. For categorical data, some papers use data transformation, but it introduces bias. This can be quite slow with large datasets

Multiple Imputations by Chained Equations (MICE)~\cite{little2002statistical}
is an iterative algorithm based on chained equations that uses an imputation model specified separately for each variable and involving the other variables as predictors. This work only considers numerical values on small data sets.

Imputations using Deep Learning (Datawig) ~\cite{datawig-cikm18,datawig-jmlr19} is a library that learns Machine Learning models using Deep Neural Networks to impute missing values. 
It also supports both CPU and GPU for training and uses feature encoder to handle categorical data.
This method works well with categorical and non-numerical features, but needs the columns as input that contain information about the target column to be be imputed. 
This is quite slow, especially with large datasets. 

NADEEF: A commodity data cleaning system~\cite{nadeef-sigmod13} allows the users to specify multiple types of data quality rules, which uniformly define what is wrong with the data and (possibly) how to repair it through writing code that implements predefined classes.
Such rule based systems achieve high precision for imputation, but this often requires a domain expert in the loop to generate and maintain the set of rules to apply. Other data imputation techniques based on eigen values include singular value decompositions and bayesian principal component analysis. The main drawback of all these techniques work well with numerical data only.

\section{Imputation Model}
\begin{algorithm}[htb]
\scriptsize
\SetAlgoLined
\SetAlgoLined\DontPrintSemicolon

 \SetKwFunction{FMain}{get\_constraints}
  \SetKwProg{Fn}{function}{:}{}
  \Fn{\FMain{data}}{ \label{func:get_constraints}
   datatypes = get\_datatypes(data) \\
   col\_constraints = get\_column\_constraints(data, datatypes) \\
   associations = get\_associations(data, datatypes) \\
  }

  \SetKwFunction{FMain}{get\_datatypes}
  \SetKwProg{Fn}{function}{:}{}
  \Fn{\FMain{column\_data}}{ \label{func:get_datatypes}
        num\_values = size(column\_data)\\
        num\_uniques\_values = size(unique(column\_data)))  \\
        \lIf {num\_values = 0} {\Return{EMPTY}}
	\lIf {has\_real\_values(column\_data)} {\Return{FLOAT}}
         \lIf {has\_date\_values(column\_data)} {\Return{DATE}}
	   \eIf {num\_unique\_values < max(log (num\_values), 20)}{
		\eIf {has\_int\_values(column\_data)} {
			return CAT\_NUM\\
		} {
			return CAT\_TEXT
		}
	}{
		\eIf {has\_int\_values(column\_data)} {
			return NUMERIC
		} {
			return TEXT
		}
	}
  }

   \SetKwFunction{FMain}{get\_column\_constraints}
  \SetKwProg{Fn}{function}{:}{}
  \Fn{\FMain{data}}{ 
  \label{func:get_column_constraints}
      \ForEach {column $c$ in data}{
	$dt$ = get\_datatype(c.data)\\
	 \uIf{$dt$ $\in$ (CAT\_NUM, CAT\_TEXT)} {
      		c.frequency = frequency\_distribution(c.values)
      	}
      	\uElseIf{$dt$ $\in$ (NUMERIC, FLOAT)} {
      		c.min = min(c.values)\\
		c.max = max(c.values)\\
		c.mean = mean(c.values)\\
		c.dist = distribution(c.values)
        }\uElseIf{$dt$ = DATE}{
        		c.mindate = mindate(c.values)\\
		c.maxdate = maxdate(c.values)\\
		c.dataformat = get\_format(c.values)
        }
     }
  }
  \caption{Constraints Inference}\label{algo:constraints}
  
\end{algorithm}

\begin{algorithm}
\scriptsize
\SetAlgoLined
\SetAlgoLined\DontPrintSemicolon
  \SetKwFunction{FMain}{get\_associations}
  \SetKwProg{Fn}{function}{:}{}
  \Fn{\FMain{data}}{ 
  \label{func:get_associations}
  	association\_list = empty \\
      \ForEach {columns $c_1,c_2$ in data}{
	association a\\
	a.source = $c_1$\\
    	a.target = $c_2$\\
      	$dt_1$ = get\_datatype($c_1$)\\
	$dt_2$ = get\_datatype($c_2$)\\
	\uIf{$dt_1 \in$ (CAT\_NUM, CAT\_CAT)} {
	 \uIf{$dt_2 \in$ (CAT\_NUM, CAT\_CAT)} {
		a.type = CAT-CAT\\
		 \ForEach {v in $c_1$}{
		 	target\_vals = get\_target\_vals(v)\\
			a.src\_value = v\\
			a.frequency = get\_freq(target\_vals)  \\
			association\_list.add(a)
		 }
		 }
     
         \uElseIf{$dt_2 \in$ (NUMERIC, FLOAT)}{
     	a.type = CAT-NUM\\
		 \ForEach {v in $c_1$}{
		 	target\_vals = get\_target\_vals(v, $c_2$) \\
			a.src\_value = v \\
			a.dist, a.error = get\_dist(target\_values)\\
			association\_list.add(a)
		 }
         }}
         \uElseIf{$dt_1 \in$ (NUMERIC, FLOAT)} {
	 \uIf{$dt_2 \in$ (NUMERIC, FLOAT)} {
	 	a.type = NUM-NUM\\
		 a.poly, a.error = get\_polynomial($c_1,c_2$)\\
		 association\_list.add(a)\\
		 \ForEach{category column $c_3$ in data}{
		 
				a.type = CAT-NUM-NUM\\
				a.catcol = $c_3$ \\
				\ForEach {v in $c_3$.data}{
					$v_1$= get\_target\_vals(v, $c_1$) \\
					$v_2$= get\_target\_vals(v, $c_2$) \\
					a.poly, a.error = get\_polynomial($v_1,v_2$)\\
					association\_list.add(a)
			}

		 }
	 }}
%
         
%
     }
     
  }

\caption{Constraints Inference - Associations}\label{algo:associations}
\end{algorithm}

\begin{algorithm}[]
\scriptsize
\SetAlgoLined
\SetAlgoLined\DontPrintSemicolon

  \SetKwFunction{FMain}{impute}
  \SetKwProg{Fn}{function}{:}{}
  \Fn{\FMain{data, constraints}}{ 
  \label{func:impute}
   Graph g = Graph(data.columns, constraints.associations)\\
  order = g.topological\_sort()\\
  \ForEach{row r in data}{
  	\ForEach {missing\_val v in order(r)} {
		$c$ = column(v)\\
		$dt$ = c.datatype \\
		\If{$dt \in$ (NUMERIC, FLOAT)}{
			v = impute\_num\_num(c, r, constraints) \\
			\lIf{val = empty}{v = impute\_cat\_num\_num(c, r, constraints)}
			\lIf{val = empty}{v = impute\_cat\_num(c, r, constraints)}
			\lIf{val = empty}{v =mean(c)}
		}
		\ElseIf{$dt \in$ (CAT\_NUM, CAT\_TEXT)}{
			v = impute\_num\_cat(c, r, constraints) \\
			\lIf{v = empty}{v = impute\_cat\_cat(c, r, constraints)}
			\lIf{v = empty}{v =most\_frequent(c)}
		}
		\ElseIf{$dt$ = TEXT}{
			v = impute\_cat\_text(c, r, constraints) \\
			\lIf{v = empty}{v =most\_frequent(c)}
		}
		\ElseIf{$dt$ = DATE}{
			v = impute\_date\_date(c, r, constraints) \\
			\lIf{v = empty}{val =mean(c)}
		}
	}
	}
  }

 \SetKwFunction{FMain}{impute\_num\_num}
  \SetKwProg{Fn}{function}{:}{}
  \Fn{\FMain{column, row, constraints}}{ 
  \label{func:num-num}
  	min\_error = $\infty$ \\
	\ForEach {association a in constraints} {
		\If{a.type = NUM-NUM $\wedge$ a.target = column}{
			\If{a.error < min\_error}{
			v = solve(a.polynomial, row(a.source)) \\
						min\_error = a.error 
			}
		}
	}
	\Return{v}
  }

 \SetKwFunction{FMain}{impute\_cat\_num}
  \SetKwProg{Fn}{function}{:}{}
  \Fn{\FMain{column, row, constraints}}{ 
  \label{func:cat-num}
  	min\_error = $\infty$ \\
	\ForEach {association a in constraints} {
		\If{a.type = CAT-NUM $\wedge$ a.target = column $\wedge$ a.src\_value = row (a.source)}{
			\If{a.error < min\_error }{
			v = get\_expected\_value(a.distribution)\\
			min\_error = a.error 
			
			}
		}
	}
	\Return{v}
  }

 \SetKwFunction{FMain}{impute\_cat\_cat}
  \SetKwProg{Fn}{function}{:}{}
  \Fn{\FMain{column, row, constraints}}{ 
  \label{func:cat-num}
  	max\_prob = $0$ \\
	\ForEach {association a in constraints} {
		\If{a.type = CAT-CAT $\wedge$ a.target = column $\wedge$ a.src\_value = row (a.source) }{
			val, prob = most\_frequent(a.frequency) \\
			\If{prob > max\_prob}{
			v = val\\
			max\_prob = prob 
			}
		}
	}
	\Return{v}
  }

   \SetKwFunction{FMain}{impute\_num\_cat}
  \SetKwProg{Fn}{function}{:}{}
  \Fn{\FMain{column, row, constraints}}{ 
  \label{func:num-cat}
	\ForEach {association a in constraints} {
		\If{a.type = CAT-NUM $\wedge$ a.source = column}{
			target\_val = row(a.target) \\
			\If {a.target.min $<$ target\_val $<$ a.target.max}{
				v.value = a.src\_value\\
				v.error = absolute(target\_val $-$ expected\_val(a.distribution))\\
				possible\_vals.add(v)	
			}
			}
			}
			vals = most\_frequent(possible\_vals)\\
			\lIf{size(vals)  > 0}{\Return{min\_error\_value(vals).value}}
			\lElse{\Return{empty}}

  } 
  
\caption{Imputation using constraints}\label{algo:impute}
\end{algorithm}

In this section, we discuss our overall solution approach spread across the subsequent subsections. The first sub-section discusses our constraints inference technique, which computes constraints from the given data. The second sub-section discusses our imputation technique using the set of inferred constraints, thereby also generating human-readable explanations for better understanding. 

\subsection{Constraints Inference}
The first step in imputing the missing values is understanding each column in the given data and finding correlations between different type of columns. We have defined seven datatypes for columns - EMPTY, DATE, TEXT, CAT\_TEXT, NUMERIC, CAT\_NUM and FLOAT. If a column does not have any value, the datatype for that column is EMPTY; if it contains date or time specific data, the datatype is DATE; if it contains string values, the datatype is TEXT or CAT\_TEXT; if it contains integer values, the datatype is NUM or CAT\_NUM; if it contains float values, the datatype is FLOAT.

Most of these datatypes are standard. The interesting and non-standard ones are CAT\_TEXT and CAT\_NUM. These datatypes cater to columns with very few unique values. For example, gender is a column with string values contains only two or three unique values while person-name is a column with string values but the number of unique values can be of the order of the number of values in the column. In order to differentiate between these two columns, we have defined separate datatypes - TEXT and CAT\_TEXT. Similarly we differentiate between NUMERIC and CAT\_NUM. This distinction helps not only in finding specific constraints at the column level, but also in finding interesting associations. For example, salary of employees may have different distributions based on the gender value.

For each column, we first find out the datatype that the column data caters to and then find the constraints based on the datatype. We have defined column level constraints for each column depending on the datatype of that column. The column level constraints for these datatypes include min, max, mean and distribution for NUMERIC, CAT\_NUM and FLOAT columns; mindate, maxdate and format for date columns; and frequency distribution for CAT\_NUM and CAT\_TEXT columns. 

In addition to constraints for each columns, we have also defined multi-column constraints, called associations, between each pair of columns depending on their datatypes. The associations are - \catcat,~\acatnum,~\acattext, ~\numnum,~\cnn~and~\datedate. All these associations, except for \cnn, describe different type of relations between two columns.
\cnn describes relations between two \num or \float columns depending on values in a categorical column. Each association has a source column and a target column. The associations are shown in Table~\ref{tab:associations}.

The \catcat association is defined for two categorical columns, and for each value in the source column, we find the frequency distribution of values in the target column.
The \catnum association is defined between a categorical column and a numerical column, and for each value in the source column, we find the min, max, mean and distribution of values in the target column. We also find error in this constraint which depicts how good fit this distribution is on the target column data. Similar to \catnum, in the \cattext association, we find frequency distribution of the target column for every value of the source column.
For the \numnum association, we try to find a polynomial function from the source column to the target column. The target column may not be an exact function of the source column, but an approximate one. So we also find the error of how good fit this polynomial is. These errors help in imputing values of a column. For example, if multiple \numnum associations are available for a target column, we choose the one with the least error. The \datedate association is a straightforward one where we find the difference between two date columns. A simple example for this association is the difference between order date and delivery date for a product.

We formally present a generalized version of the above algorithm in Algorithm~\ref{algo:constraints} and Algorithm~\ref{algo:associations}.

\begin{table}
    \centering
\begin{tabular}{ |p{1.7cm}|p{1.6cm}|p{1.6cm}|p{2cm}| } 
\hline
Association & Source  & Target  & Constraints \\
 \hline
\catcat & \cattext/ \catnum & \cattext/ \catnum & Frequency distribution \\
 \hline
 \acatnum & \cattext/ \catnum & \num/ \float  & For each source value, min, max, mean and distribution of target column \\ 
 \hline
 \acattext & \cattext/ \catnum & \ctext & Frequency distribution \\ 
 \hline
 \numnum & \num/ \float & \num/ \float & Polynomial function \\ 
 \hline
 \cnn & \num/ \float & \num/ \float & Polynomial function, for each category value \\ 
 \hline
 \datedate & \cdate & \cdate & mindiff, maxdiff \\ 
 \hline
\end{tabular}
\caption{Associations}
\label{tab:associations}
\end{table}

\subsection{Imputation using constraints}

The imputation of a missing value is done using the constraints and the values in other columns. 
The idea is to impute the values of a column using associations first, and 
if it is not possible to use associations, impute the value using column level constraints.
Note that this can happen either due to non-availability of the required associations or due to the non-availability of the values of other columns required by an association.
We will discuss the imputation algorithm informally below, and a generalized version of the  algorithm is presented formally in Algorithm~\ref{algo:impute}.

If there are multiple values missing in the same row, the values are imputed in a particular order.
This is done by constructing a graph based on the associations and then sorting it topologically. The intuition behind this is that the columns with very few unique values, for example labels in a training dataset, have more valuable information than a column with many values. Thus, the objective is to give preference to categorical columns over numerical or text columns.
Not that this is also reflected in the way associations are defined; the source columns for most associations are categorical(\acattext/\acatnum). 

For imputing a value in a categorical column, the \acatnum associations are used first and if it fails, \catcat associations are used and if that also fails, most frequent value of the column is chosen. For the imputation using \acatnum association, 
all the possible values for the missing column are computed that conform with the values in other numerical columns range specified in the constraints.
The value that conforms with most of the values in numerical columns is chosen. If there are multiple possible values that conform with most numerical columns, the one closest to the mean of the numerical columns is chosen. Similarly for \catcat association, all the possible values are computed that are most frequent for values in categorical columns, and the one with highest probability is chosen.

For imputing a numerical column, the \numnum association is used for imputing the missing value. And if it fails, i.e., if there is no \numnum association for the missing value column, or the source column value is missing, \cnn association is used. If \cnn association also fails, then \catnum association is used and if that also fails, mean value of the column is chosen.
The imputation using \numnum association is straightforward. If there are multiple \numnum associations in the constraints, the one with least error is used for imputing the value. The imputation using \cnn is similar to imputation using \numnum, subject to value in a categorical column. The imputation using \acatnum, similar to the imputation of a categorical column using \catcat association, imputes the value with the expected value of the given distribution with least error.

For imputing a text column, the \acattext association is used for imputing the missing value. And if it fails, the most frequent value of the column is used. The imputation of a \cdate column uses \datedate association if there are other date columns, otherwise the value is imputed with the median of the column.

\subsubsection*{Explanations for Imputations}
The explanations for an imputation comes directly from the constraints used. For example, if \acatnum is used for imputing a numerical value, the value and the name of the categorical column used is the explanation for the imputation. Similarly, for an imputation using \numnum constraint, the value and the name of the numerical column used is the explanation.

\section{Experimental Evaluation}
\label{sec:expt}

\begin{figure*}
\centering
\subcaptionbox{Polynomials}{\includegraphics[width=0.30\textwidth]{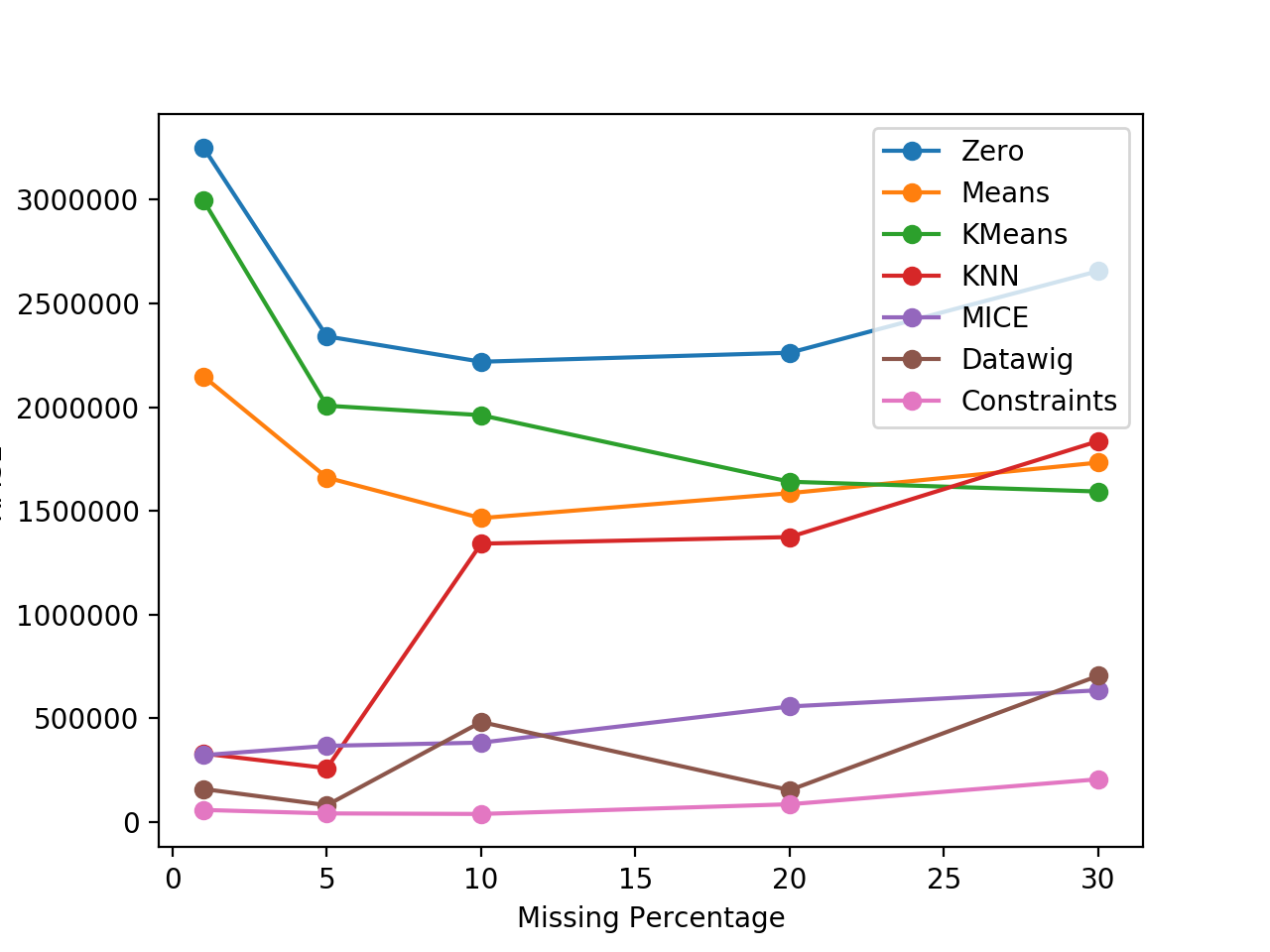}}%
\subcaptionbox{Iris}{\includegraphics[width=0.30\textwidth]{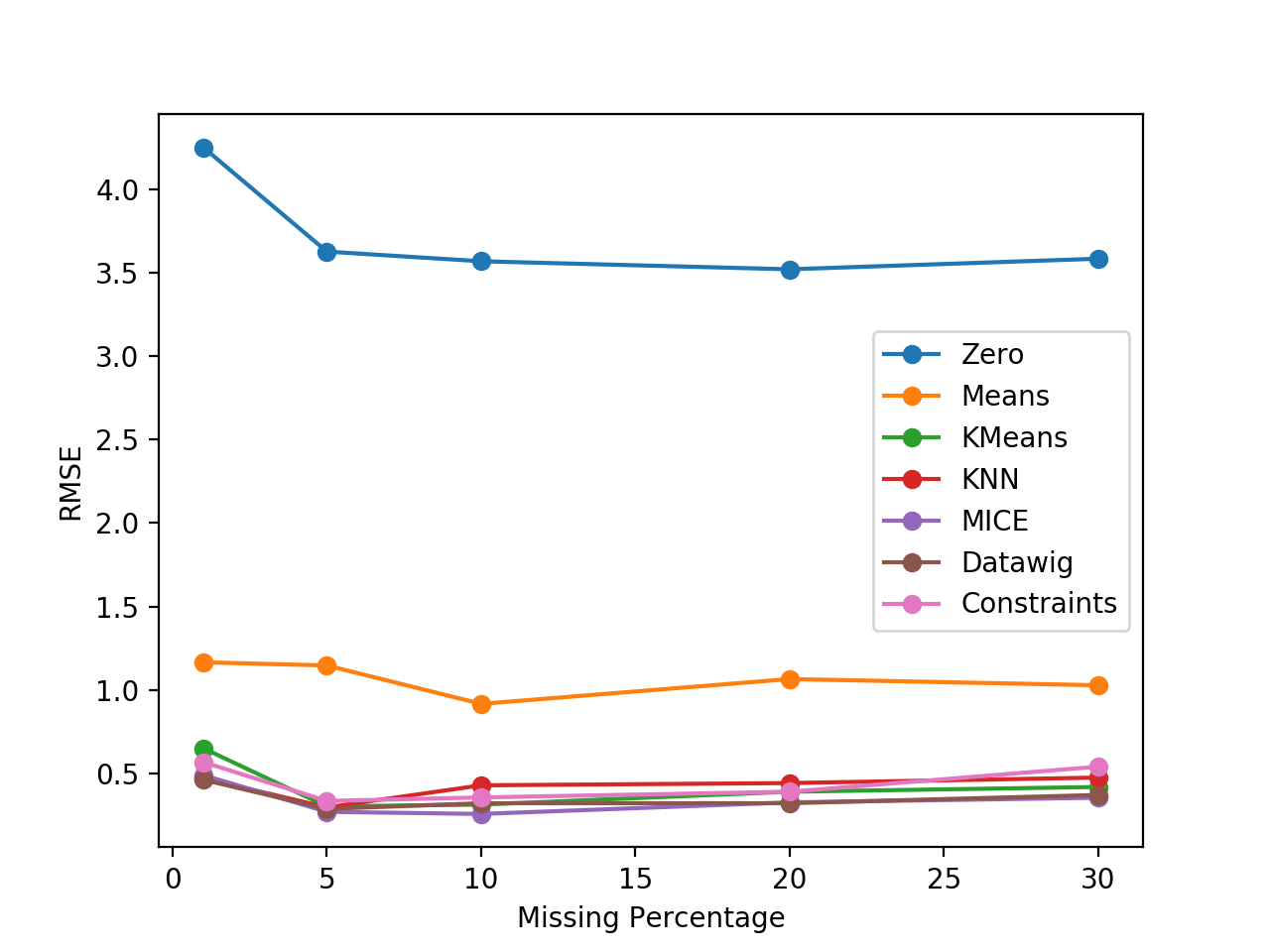}}%
\subcaptionbox{Ecoli}{\includegraphics[width=0.30\textwidth]{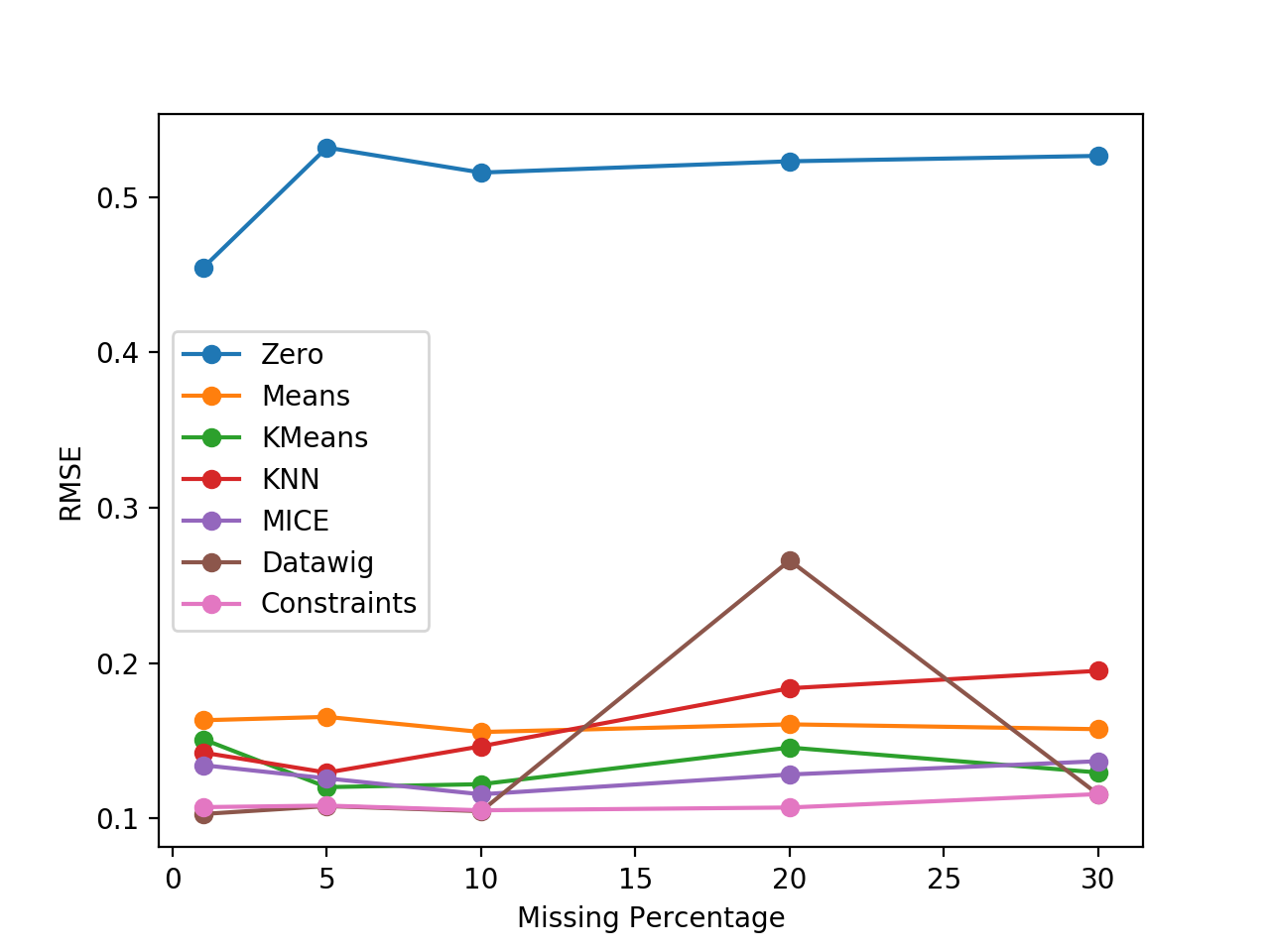}}%
\par
\subcaptionbox{Breast Cancer}{\includegraphics[width=0.30\textwidth]{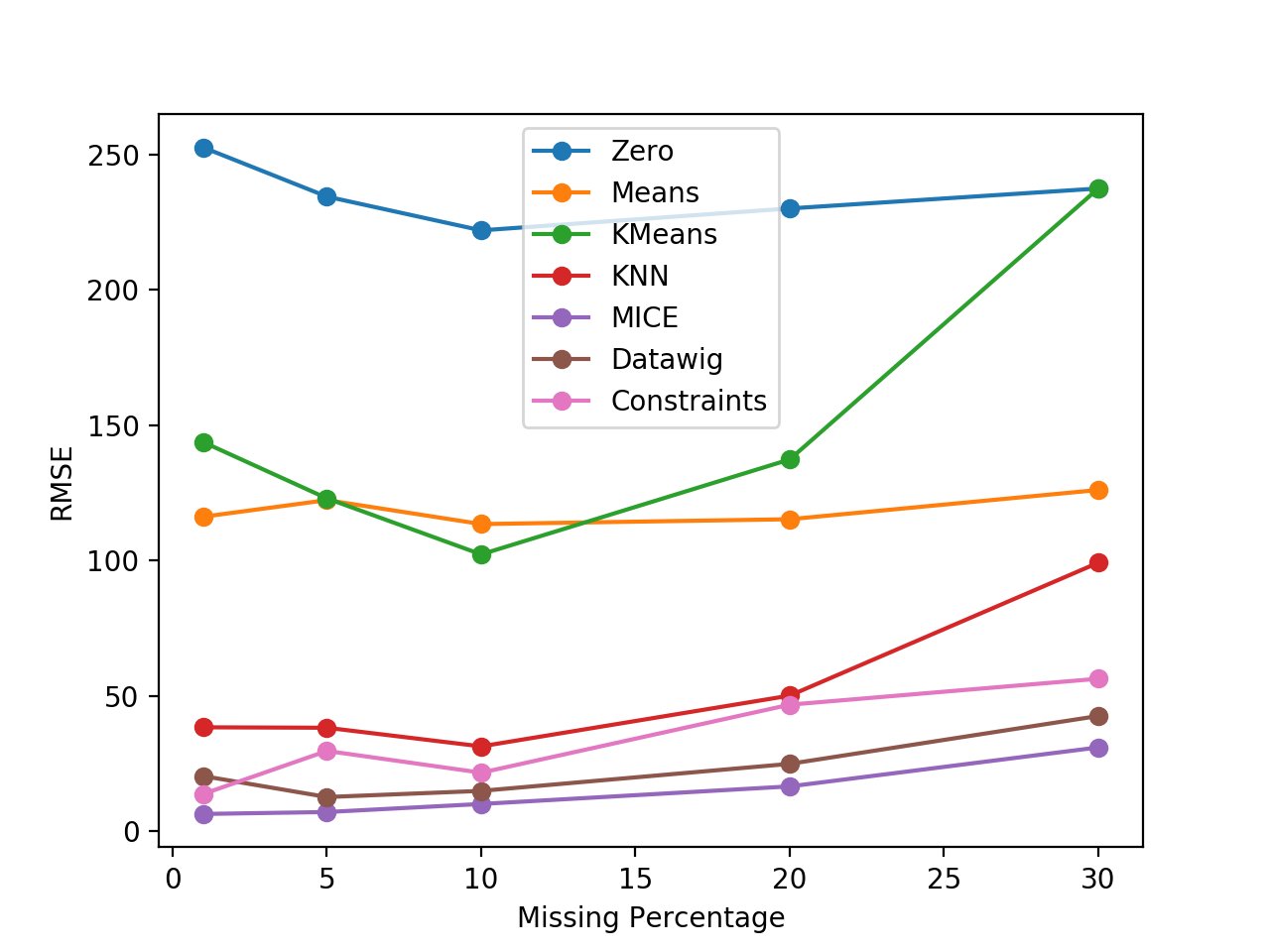}}%
\subcaptionbox{Wine}{\includegraphics[width=0.30\textwidth]{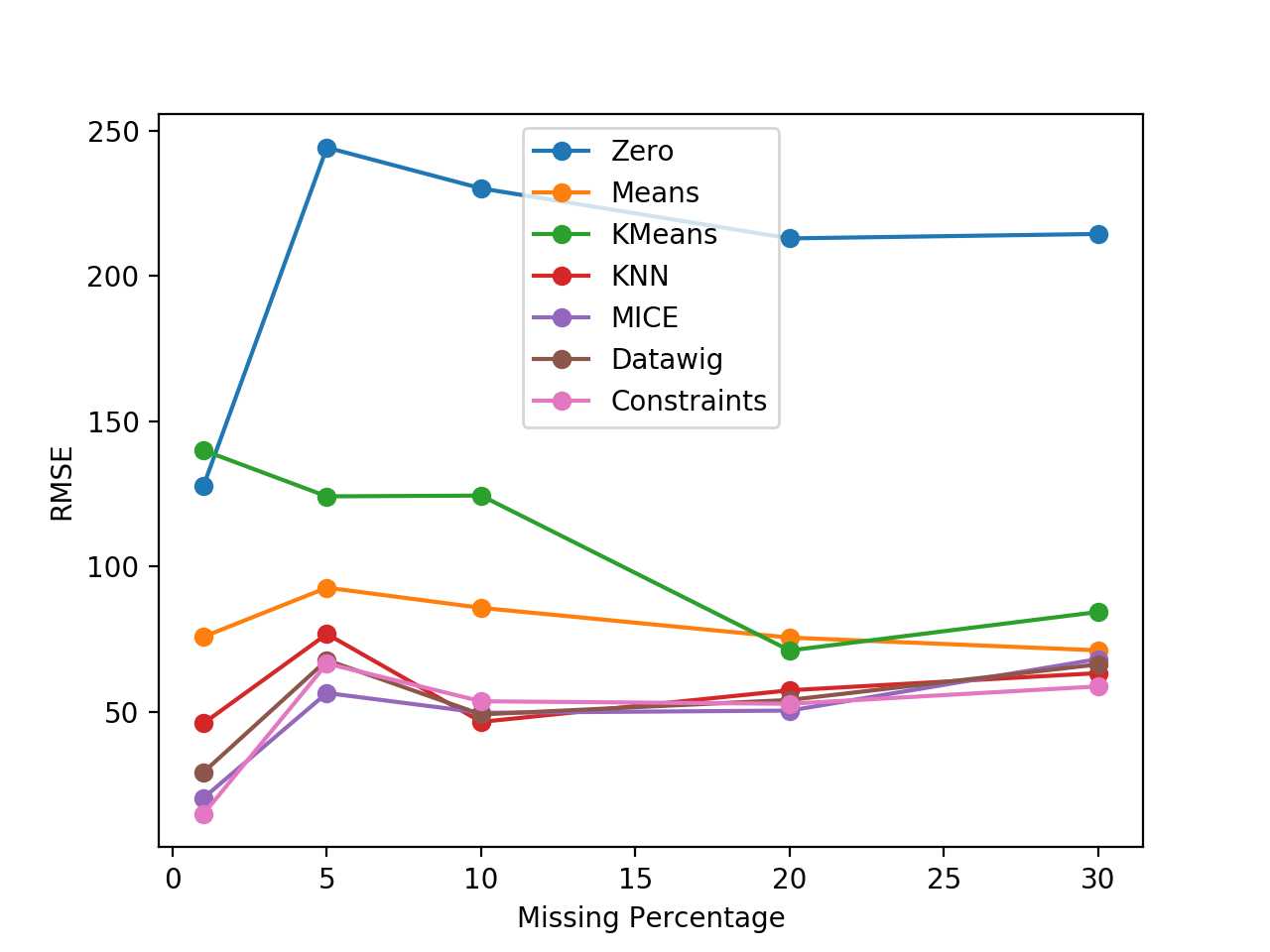}}%
\subcaptionbox{Diabetes}{\includegraphics[width=0.30\textwidth]{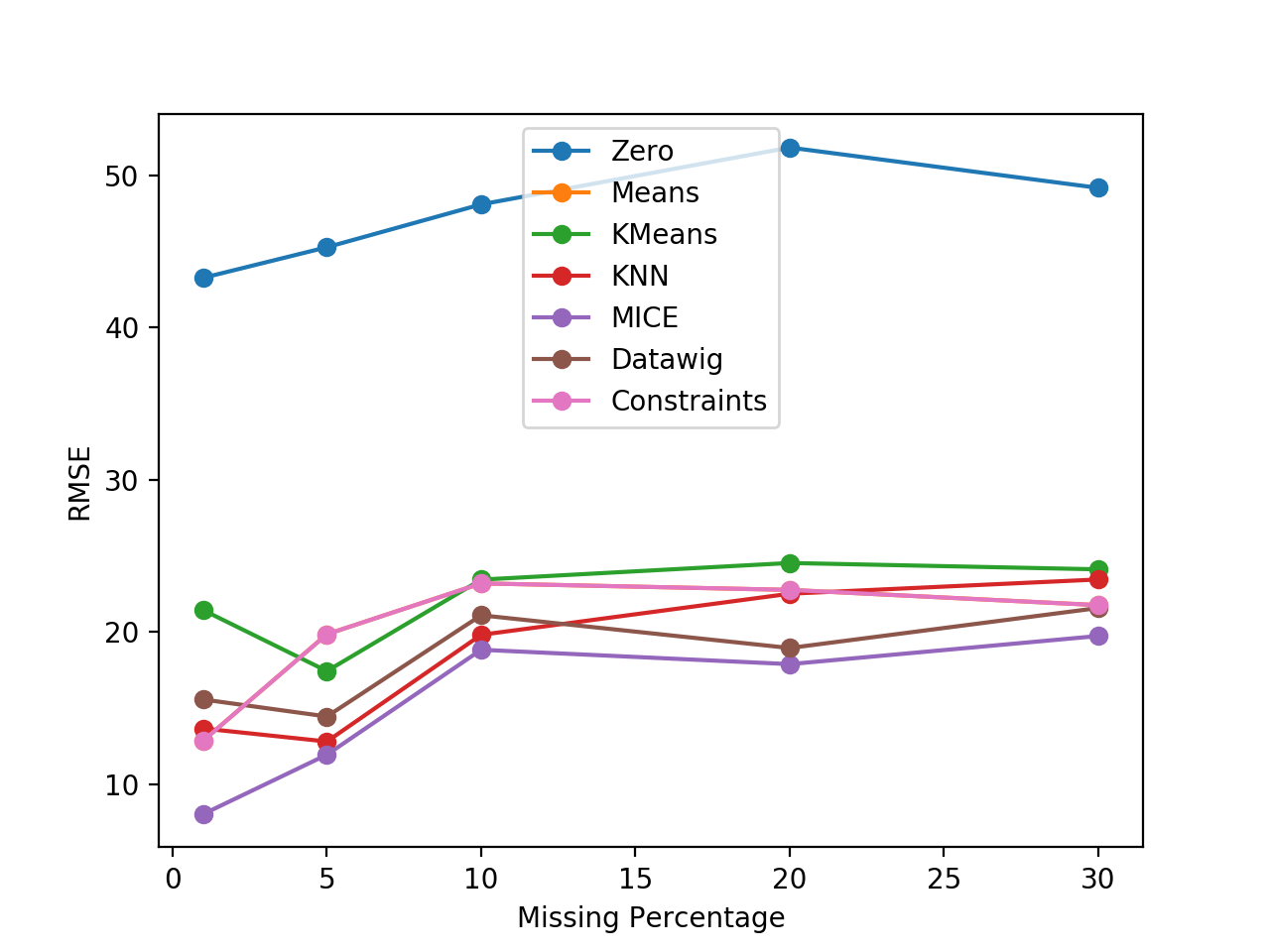}}%
\caption{RMSE for Numerical Imputation}
\label{fig:num_results}
\end{figure*}

\begin{figure*}
\centering
\subcaptionbox{Iris}{\includegraphics[width=0.30\textwidth]{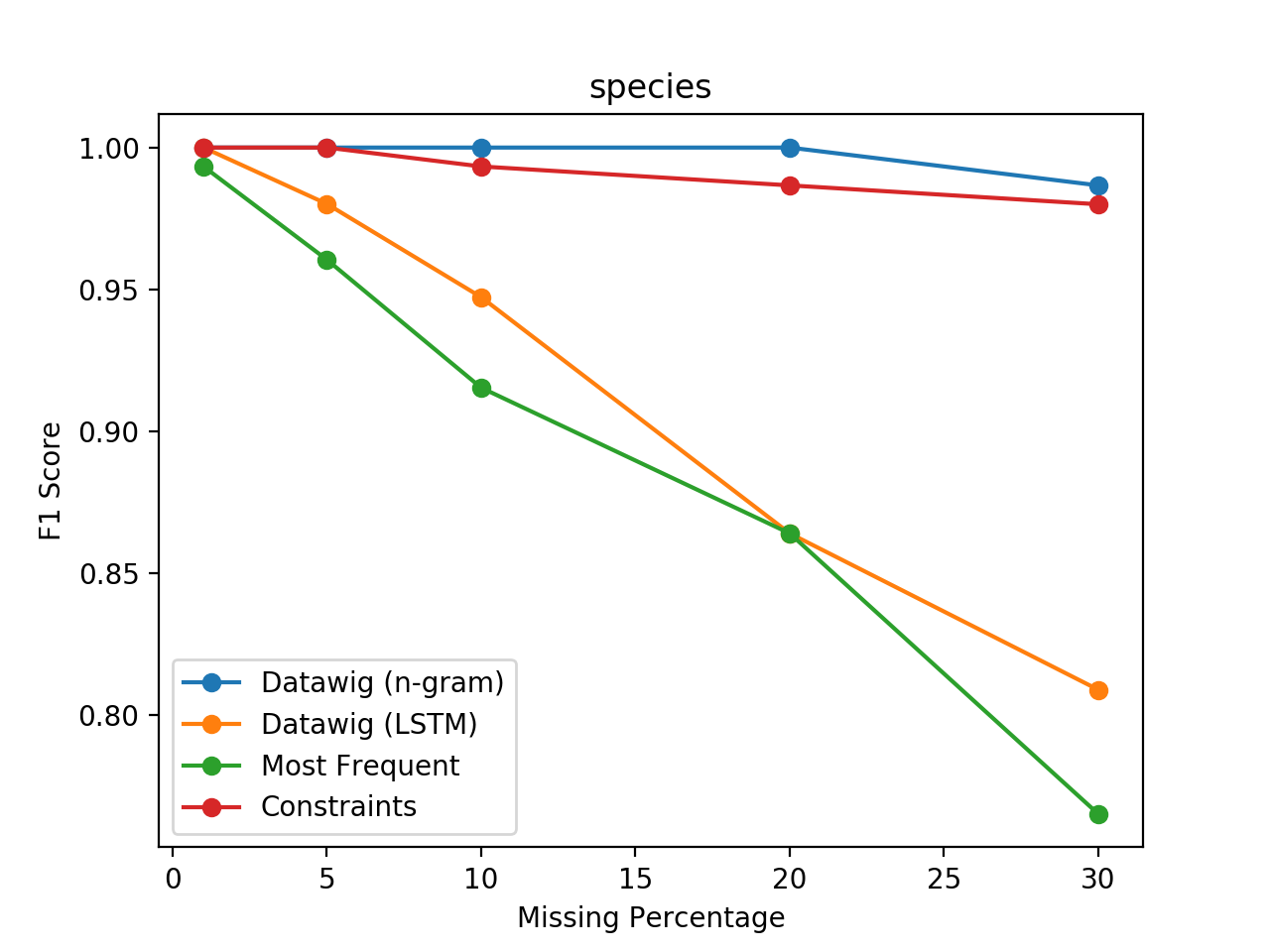}}%
\subcaptionbox{Bank Market(Job)}{\includegraphics[width=0.30\textwidth]{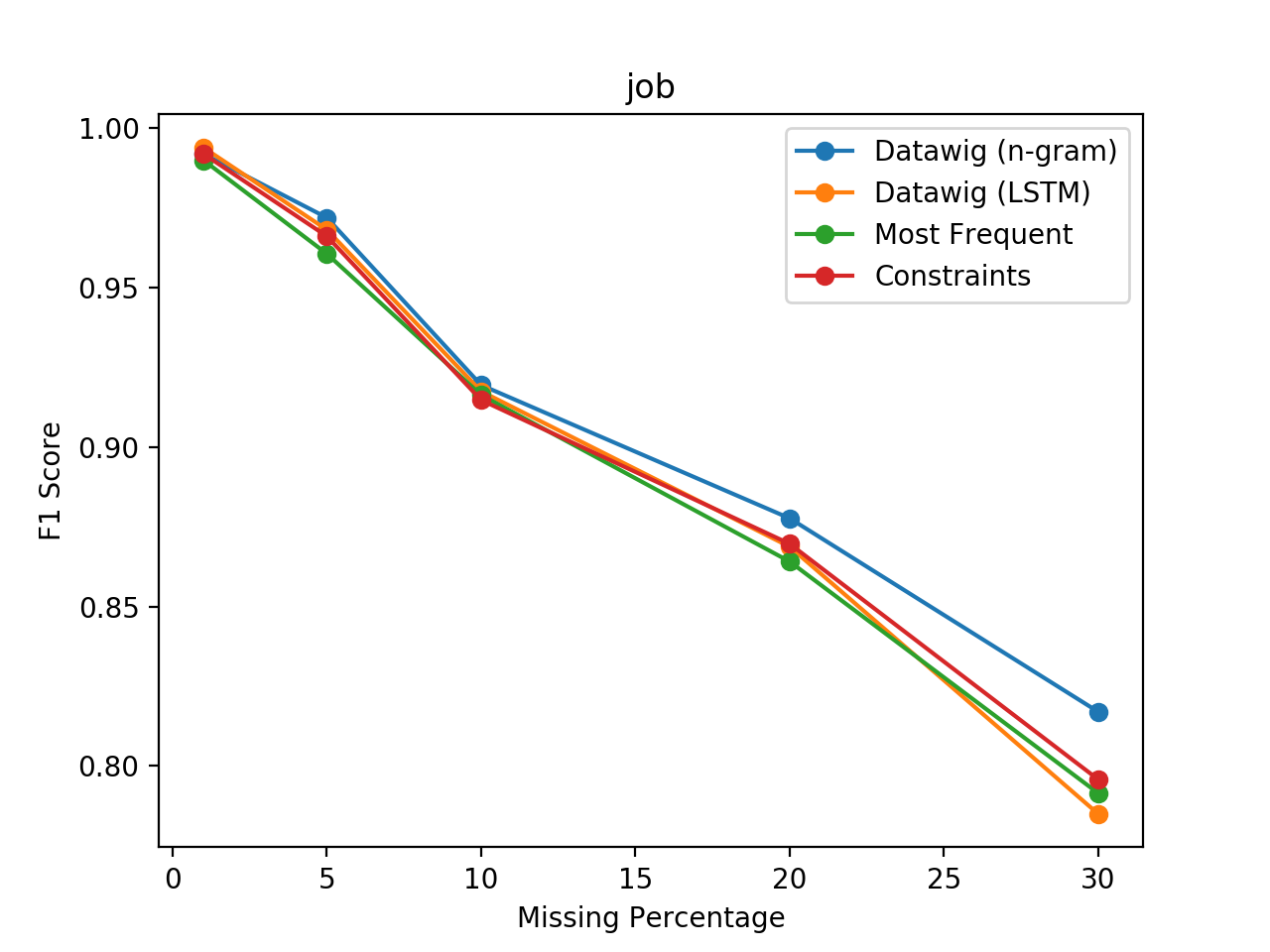}}%
\par
\subcaptionbox{Bank Market(Marital)}{\includegraphics[width=0.30\textwidth]{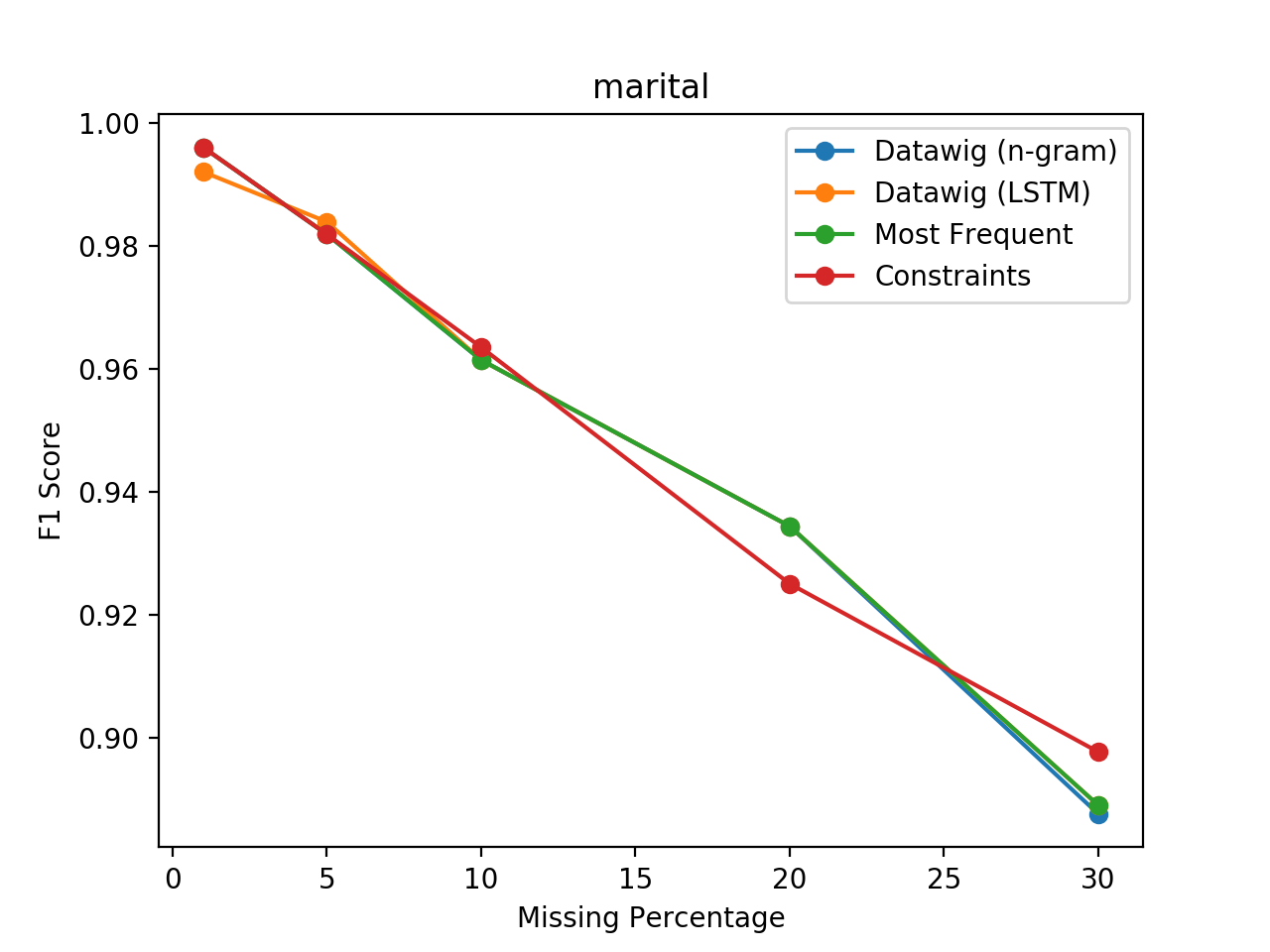}}%
\subcaptionbox{Bank Market(education)}{\includegraphics[width=0.30\textwidth]{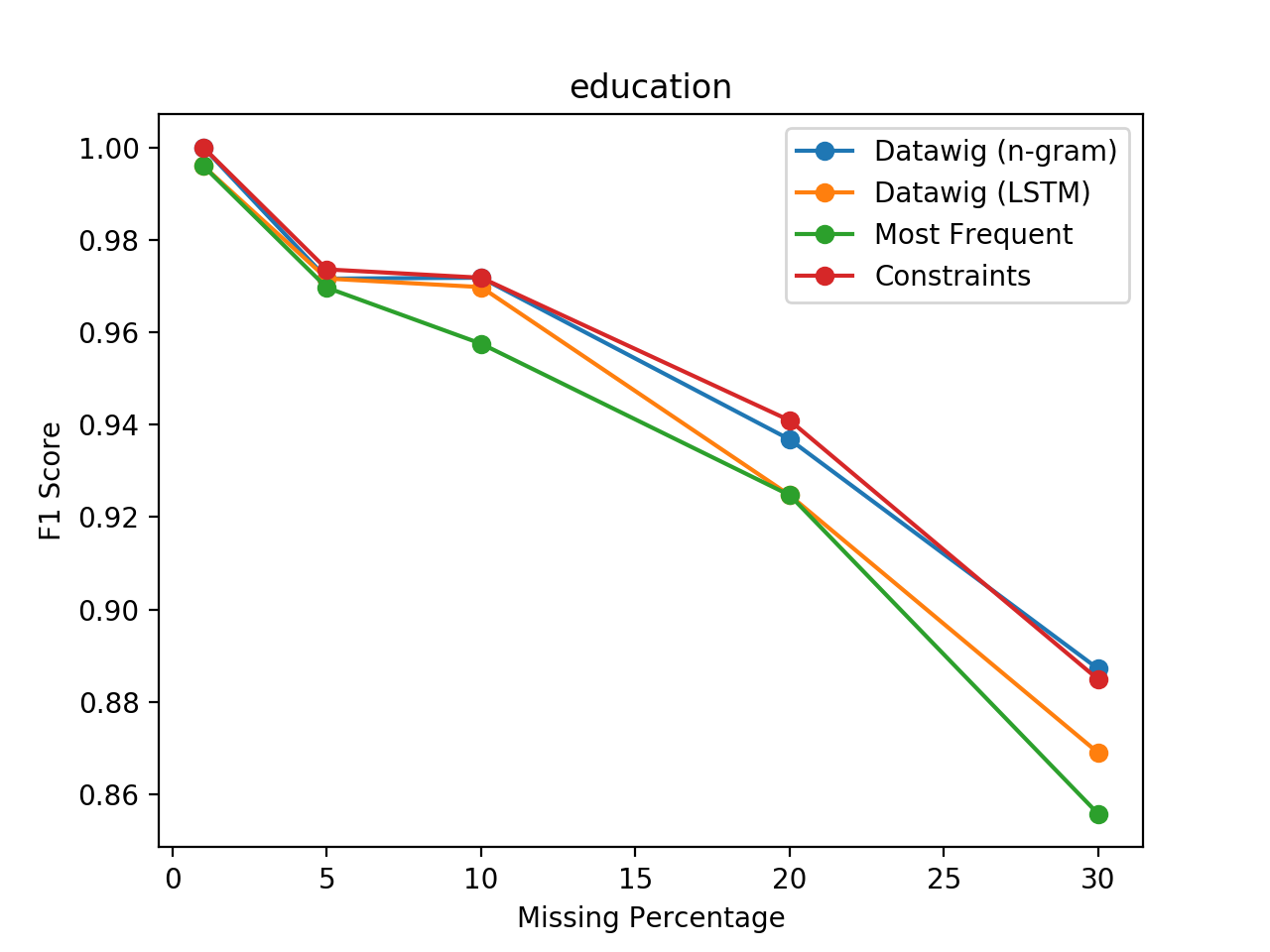}}%
\caption{F1 score for Categorical Imputation}
\label{fig:cat_results}
\end{figure*}

\subsection{Setup}

\begin{table}[b]
\centering
\caption{Benchmark Characteristics\label{tab:bench}}
\begin{tabular}{|c|c|c|c|}
\hline
Benchmark & Size & \#Features \\
\hline
Polynomial $^1$& 1000 & 5 \\
Iris $^2$& 150 & 4 \\
Ecoli $^2$& 336 & 8 \\
Wine $^2$& 178 & 13 \\
Diabetes $^2$& 486 & 20 \\
Breast Cancer $^2$& 286 & 9 \\
\hline
\end{tabular}
\end{table}

\begin{table}[b]
\scriptsize
\begin{tabular}{ll}\hline
$1$ & synthetic data with columns having polynomial relationships\\
$2$ & \url{https://archive.ics.uci.edu/ml/datasets.php}\\
\end{tabular}
\end{table}
\subsubsection{Benchmark Characteristics} We have assessed the performance of our approach on 
open-source data sets from varied sources as listed in Table~\ref{tab:bench}. Additionally, we also consider a synthetic data set with three numeric attributes having 
NUM-NUM polynomialassociation constraints between them.

\subsubsection{Configurations}
Our code is written in Python and executed in Python 3.7.
All the experiments are performed in a machine running macOS 10.14, having 16GB RAM, 2.7Ghz CPU running Intel Core i7.

\subsubsection{Missing data Generation}
The previously mentioned data sets which we have considered for our experiments have no missing value in their original forms. Therefore, we take a random approach to pick indices in any data set to discard their values and treat them as missing ones. We consider a missing data percentage variable, \emph{perc.} to define how many values in the entire data set are treated as missing.
It is further to be noted that we first encode the categorical text values present in the data sets using an appropriate encoder before feeding it to the data imputation engine. Such values which were treated as CAT\_TEXT in their original form are now marked with a datatype CAT\_NUM by our constraint inference module.
\subsubsection{Experiment Runs}
We take an iterative approach while running our experiments to augment the reliability and correctness of our results. We have set the variable \emph{iter} as $5$ for all our experimental runs, which means that 5 consecutive imputation rounds were performed to replace the missing values in an input data set. The numbers reported for different metrics in the subsequent subsections are the average of all the iterations in a single experiment run.

\subsection{Experiment Goal}
We have planned our set of experiments in an attempt to find out how well does our approach perform as compared to the already existing ones. We have considered three different metrics for comparison, namely \emph{Data Accuracy}, \emph{Prediction Accuracy} and \emph{Fidelity}. Please note that most of the prior works have reported only Data Accuracy. But, we are evaluating on two additional metrics to further assess how the missing value imputation using different approaches impact the model accuracy. A detailed description of these metrics is presented along with the comparative evaluation of different approaches in the next Subsection \ref{compare}.


\subsection{Comparison with the Related Works}
\label{compare}
We have compared the performance of our approach to fill missing gaps in data  as compared to the existing ones, such as mean, k-means, KNN~\cite{knn-aai03}, MICE~\cite{little2002statistical}, MissForest~\cite{missforest-bioinformatics12,van2018flexible}, Datawig~\cite{datawig-cikm18,datawig-jmlr19}. We leverage the existing functionalities in sklearn to implement some of these prior works. The implementations for mean, KNN have been taken from \emph{fancyimpute} package, and \emph{IterativeImputer} with the estimators \emph{RandomForestRegressor} and \emph{LinearRegression} caters to mimic the MissForest and MICE, respectively. For k-means, we use sklearn's implementation with cluster count set as 4.
Further, the freely available python-based Datawig API \footnote{https://pypi.org/project/datawig/} is used to fetch performance numbers for Datawig.

Next, we discuss our different evaluation  metrics along with the obtained experimental results.

\begin{table}[t]
\caption{$NRMSE$ for ours vs related works}
\centering
\begin{tabular}{|l|l|l|l|l|l|l|l|l|}
\hline
Bench. & perc. & \multicolumn{6}{|c|}{Avg. $NRMSE$ for $iter$=$5$} \\ \cline{3-8} 
&  & mean & k-means & KNN & MICE & Datawig & Ours \\ \hline
Wine & 5 &  0.99 & 0.99 & 1.01 & 0.62 & 0.65 & 0.70  \\ \cline{2-8} 
& 10 &  0.98 & 0.99 & 0.93 & 0.66 & 0.68 & 0.69 \\ \cline{2-8} 
& 20 &  0.99 & 1.11 & 0.87 & 0.68 & 0.75 & 0.70   \\ \cline{2-8} 
& 30 & 0.98 & 2.03 & 0.88 & 0.82 & 0.78 & 0.74  \\ \hline
Ecoli & 5 & 0.90 & 0.70 & 0.61 & 0.68 & 0.62 & 0.61  \\ \cline{2-8} 
& 10 &  0.81 & 0.61 & 0.73 & 0.63 & 0.67 & 0.59 \\ \cline{2-8} 
& 20  & 0.84 & 0.66 & 0.90 & 0.76 & 0.70 & 0.63 \\ \cline{2-8} 
& 30 & 0.92 & 0.85 & 1.04 & 1.04 & 0.84 & 0.80  \\ \hline

Polynomials & 5 & 3.87 & 4.69 & 2.82 & 1.08 & 1.34 & 0.90  \\ \cline{2-8} 
& 10 & 4.40 & 5.23 & 3.79 & 1.27 & 1.11 & 0.87  \\ \cline{2-8} 
& 20  & 4.20 & 4.5 & 4.19 & 1.45 & 1.25 &  1.59  \\ \cline{2-8} 
& 30 & 4.18 & 4.98 & 4.54 & 1.98 & 1.65 & 2.45  \\ \hline

\end{tabular}
\label{nrmse}
\end{table}


  
  
  
 
  


\textbf{RMSE and F1 Score.} For different benchmarks, using different approaches, we record the RMSE values obtained for the numerically imputed versions in Figure \ref{fig:num_results}. For the categorical imputations, we record F1-score as shown in Figure \ref{fig:cat_results}. The plots in the two sets of figures clearly shows that our technique imputes data better than the existing techniques by introducing less outliers in the imputed versions. 

\textbf{NRMSE.} For a column $c$ with datatype NUMERIC or CAT\_NUM, the normalized root mean square error with a standard deviation $\sigma_c$, denoted by $NRMSE_c$, is computed as follows: 
\[
NRMSE_c = \frac{RMSE_c}{\sigma_c}
\]

Hence, the normalized root mean square for a dataset, with $NCOL$ as the set of all NUMERIC or CAT\_NUM columns, is the mean of normalized root mean square error for all such columns.
\[
NRMSE = \frac{\displaystyle\sum_c^{NCOL}{NRMSE_c}}{|NCOL|} 
\]

The NRMSE for different benchmarks with different missing values percentage resulted using different imputation approaches is listed in Table \ref{nrmse}. Please note that the values reported here are the average ones across $5$ consecutive iterations.


\textbf{Prediction Accuracy and Fidelity.} 
The experiment starts with splitting the input dataset as $train\_inputs$ and $test\_inputs$ in 70:30 respectively. We then train a Decision Tree Classifier or Regressor depending on the datatype of class label using $train\_inputs$, and record the accuracy on the $test\_inputs$. This accuracy acts as the baseline for future comparison for this particular dataset.

\begin{table}[t]
\caption{$Prediction$ $Accuracy$ and $Fidelity$}
\centering
\begin{tabular}{|l|l|l|l|l|l|l|l|l|}
\hline
Bench. & perc. & \multicolumn{6}{|c||}{($Prediction$ $Accuracy$, $Fidelity$) for $iter$=$5$} \\ \cline{3-8} 
&  & mean & k-means & KNN & MICE & Datawig & Ours\\ \hline

Wine & 5 & 0.88, 0.81 & 0.88, 0.81 & 0.87, 0.79 & 0.87, 0.88  & 0.90, 0.92 & 0.98, 0.85\\ \cline{2-8} 
& 10 & 0.85, 0.74 & 0.77, 0.72 & 0.83, 0.75  & 0.81, 0.79  & 0.88, 0.85 & 0.98, 0.85  \\ \cline{2-8} 
& 20  & 0.79, 0.72 & 0.83, 0.81 & 0.79, 0.77 &  0.88, 0.79& 0.85, 0.77 & 1, 0.87 \\ \cline{2-8} 
& 30 & 0.83, 0.74  & 0.64, 0.62 & 0.81, 0.75 & 0.75, 0.77  & 0.83, 0.79 & 0.92, 0.79   \\ \hline

Iris & 5 & 0.95, 0.95  & 0.95, 0.95 & 1, 1 & 0.97, 0.97 & 0.97, 0.97 & 1, 1  \\ \cline{2-8}  
& 10 & 0.84, 0.84 & 0.95, 0.95 & 1, 1  &  0.95, 0.95 & 1, 1 & 1, 1  \\ \cline{2-8} 
& 20 & 0.75, 0.75 & 0.91, 0.91 & 0.91, 0.91 & 0.95, 0.95  & 0.91, 0.91 & 0.95, 0.95   \\ \cline{2-8} 
& 30 & 0.77, 0.77 & 0.91, 0.91  & 0.86, 0.86 & 0.93, 0.93 & 0.95, 0.93 & 0.95, 0.95  \\ \hline
\end{tabular}
\label{tab:accuracy}
\end{table}
\begin{figure}
\centering
\begin{subfigure}{.5\textwidth}
  \centering
  \includegraphics[width=0.5\textwidth]{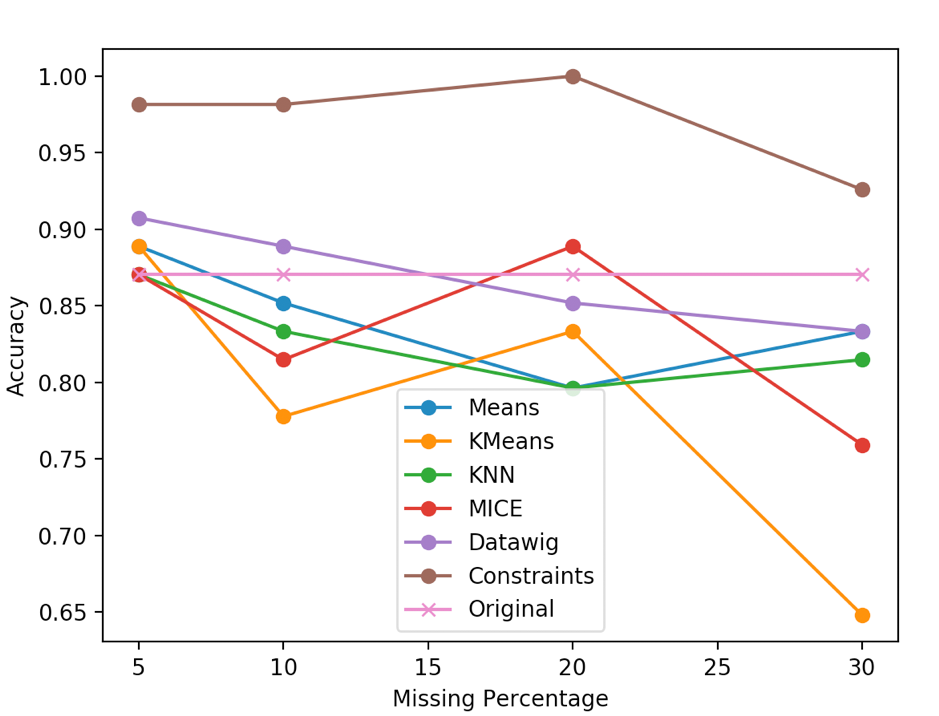}%
\includegraphics[width=0.5\textwidth]{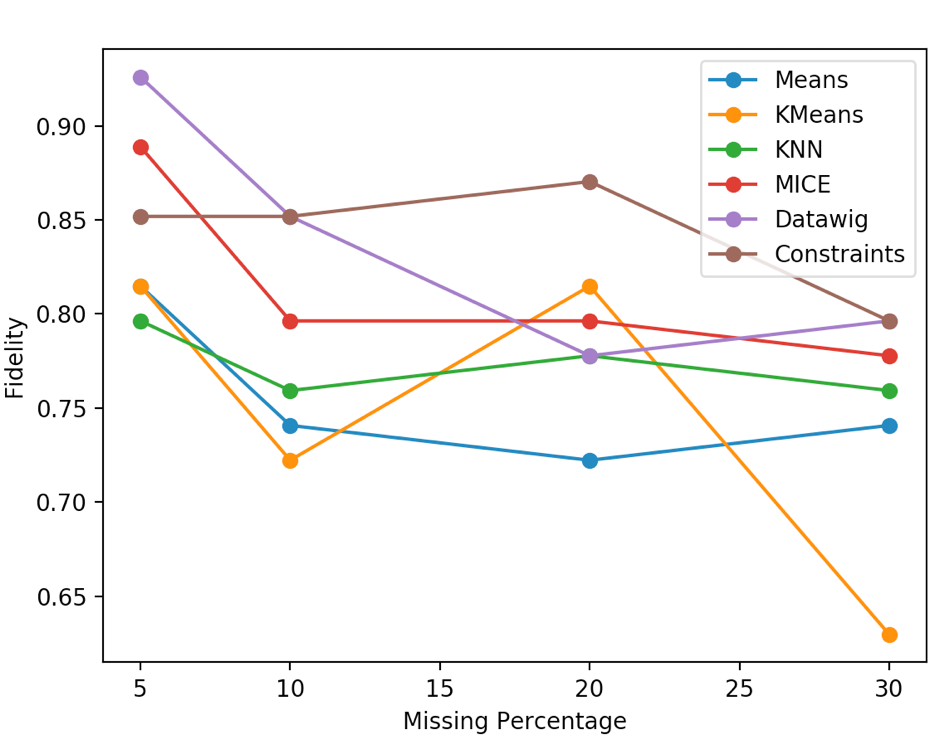}

    \caption{Wine Dataset}

  \label{fig:sub-second}
\end{subfigure}


\begin{subfigure}{.5\textwidth}
  \centering
  \includegraphics[width=0.5\textwidth]{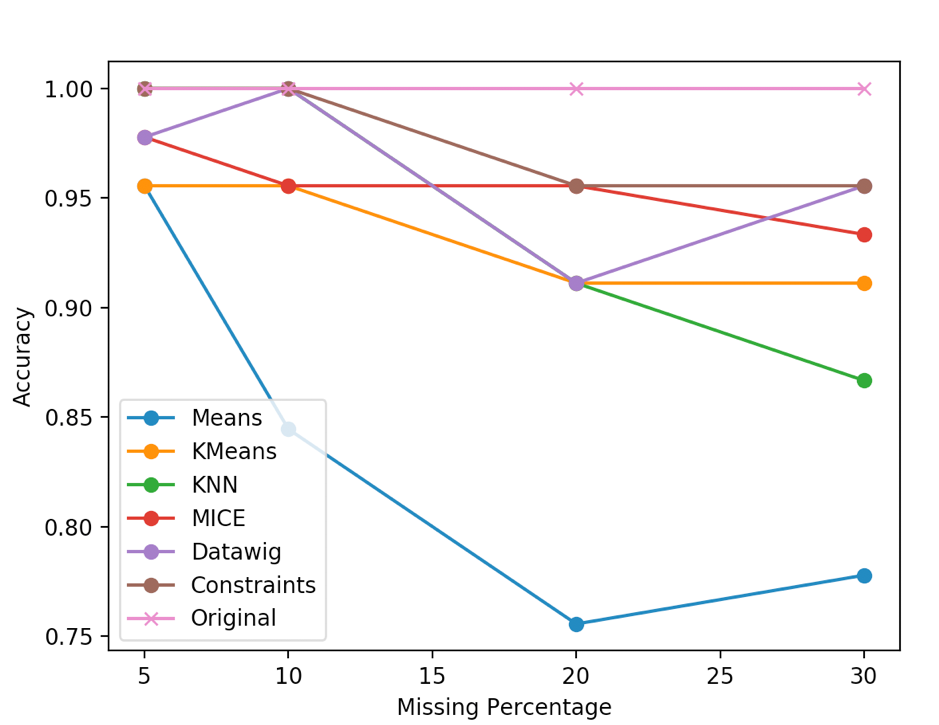}%
\includegraphics[width=0.5\textwidth]{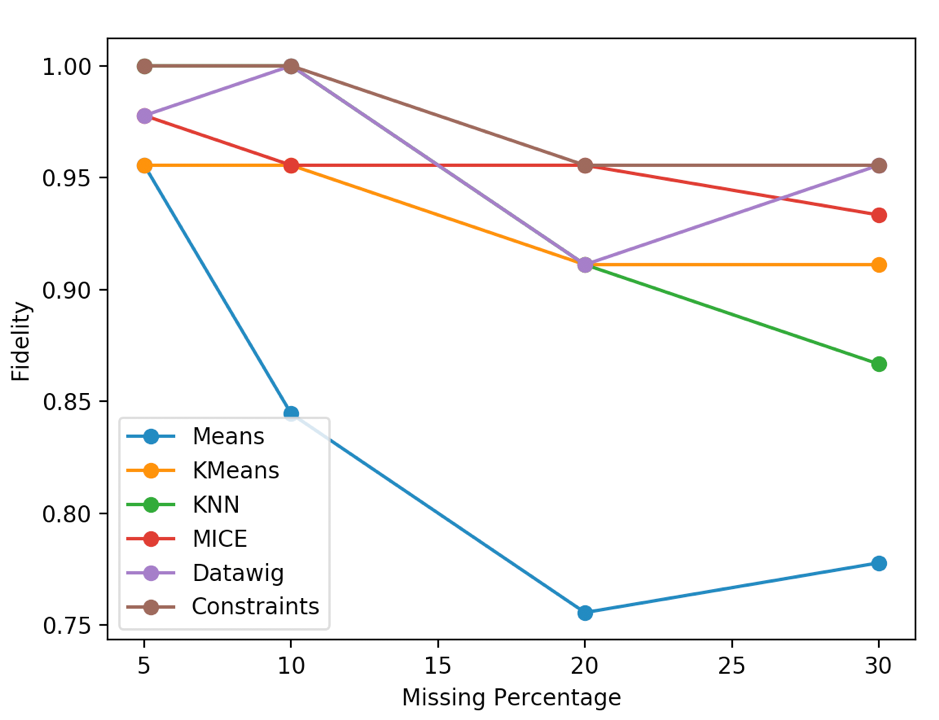}%
  \caption{Iris Dataset}
  \label{fig:sub-third}
\end{subfigure}
\caption{Accuracy and Fidelity}
\label{fig:accu_fid_plot}
\end{figure}
Next, using $train\_inputs$, we generate different versions of training inputs with missing values while varying the missing data percentage. For every generated training input set with missing value, say, $train\_impute$ we perform imputation using different state-of-the-art approaches along with our novel method. We also train a new Decision Tree Classifier or Regressor using every $train\_impute$ and record its accuracy on $test\_inputs$. Additionally, we also record the fidelity of the trained models using imputed trained data for the datasets having categorical class label. Table \ref{tab:accuracy} reports the prediction accuracy and fidelity of the models trained using imputed training data using different methods for Wine and Iris datasets. Please note that both of these datasets have categorical class label, hence, fidelity scoring was possible. As inferred from Figure \ref{fig:accu_fid_plot}, our approach performs better than the state-of-the-art methods while offering higher accuracy without compromising much on fidelity.
\\
\begin{figure}
\begin{subfigure}{.5\linewidth}
  \centering
\includegraphics[width=\linewidth]{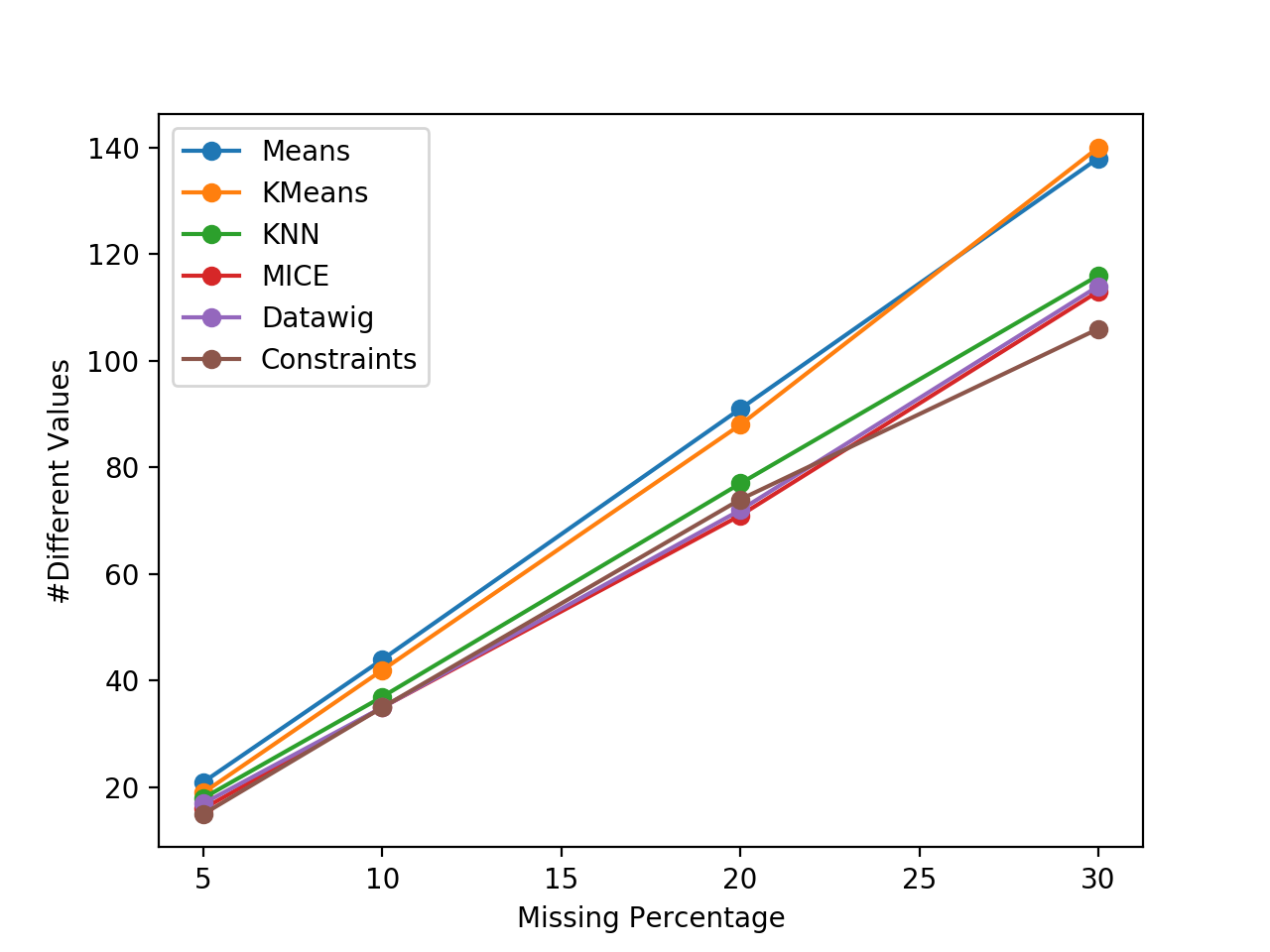}
  \label{fig:sub-first}
   \caption{Wine Dataset}
\end{subfigure}%
\begin{subfigure}{.5\linewidth}
  \centering
\includegraphics[width=\linewidth]{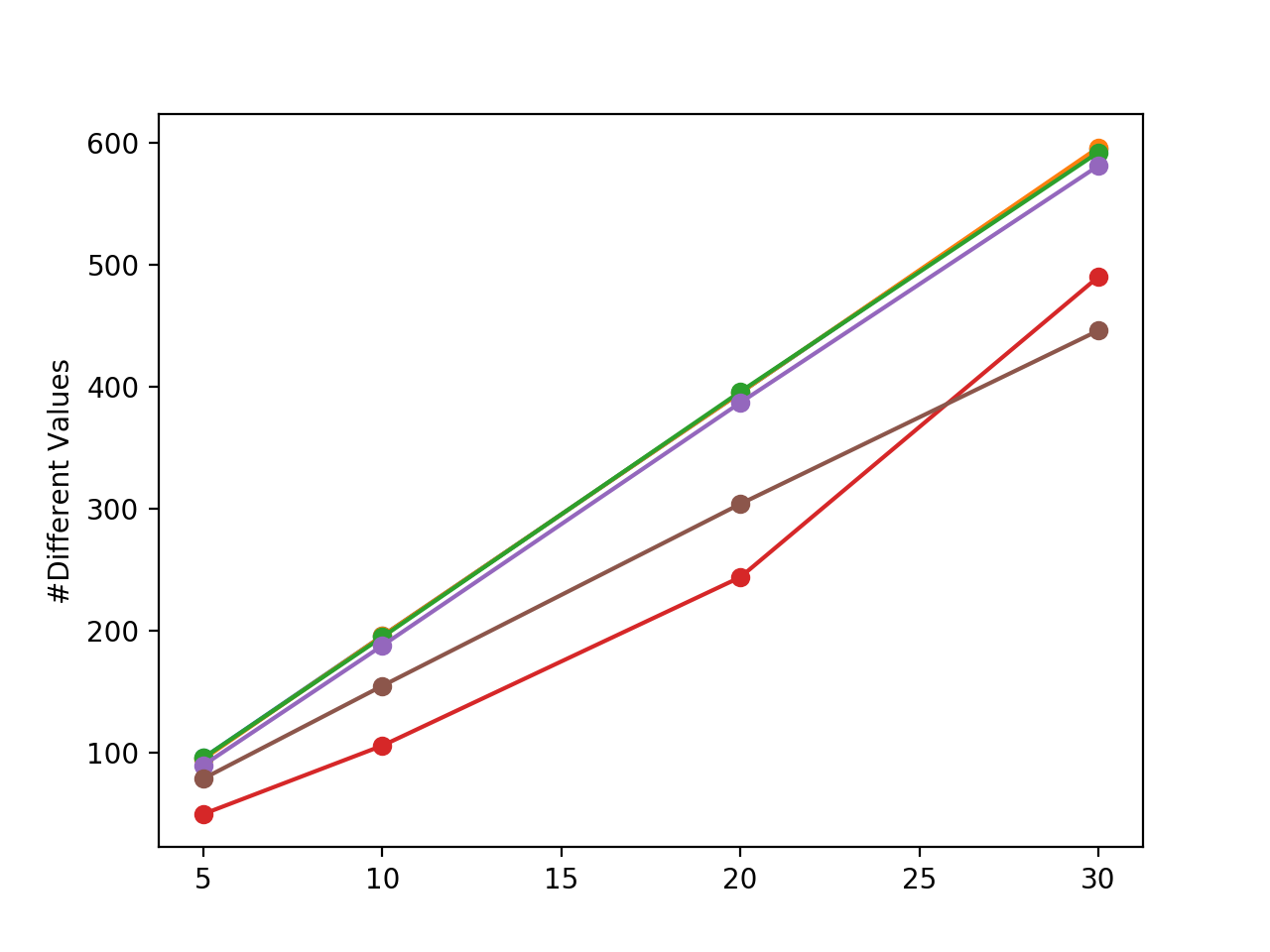}
    \caption{Polynomial Dataset}
  \label{fig:sub-second}
\end{subfigure}

\begin{subfigure}{.5\linewidth}
  \centering
  \includegraphics[width=\textwidth]{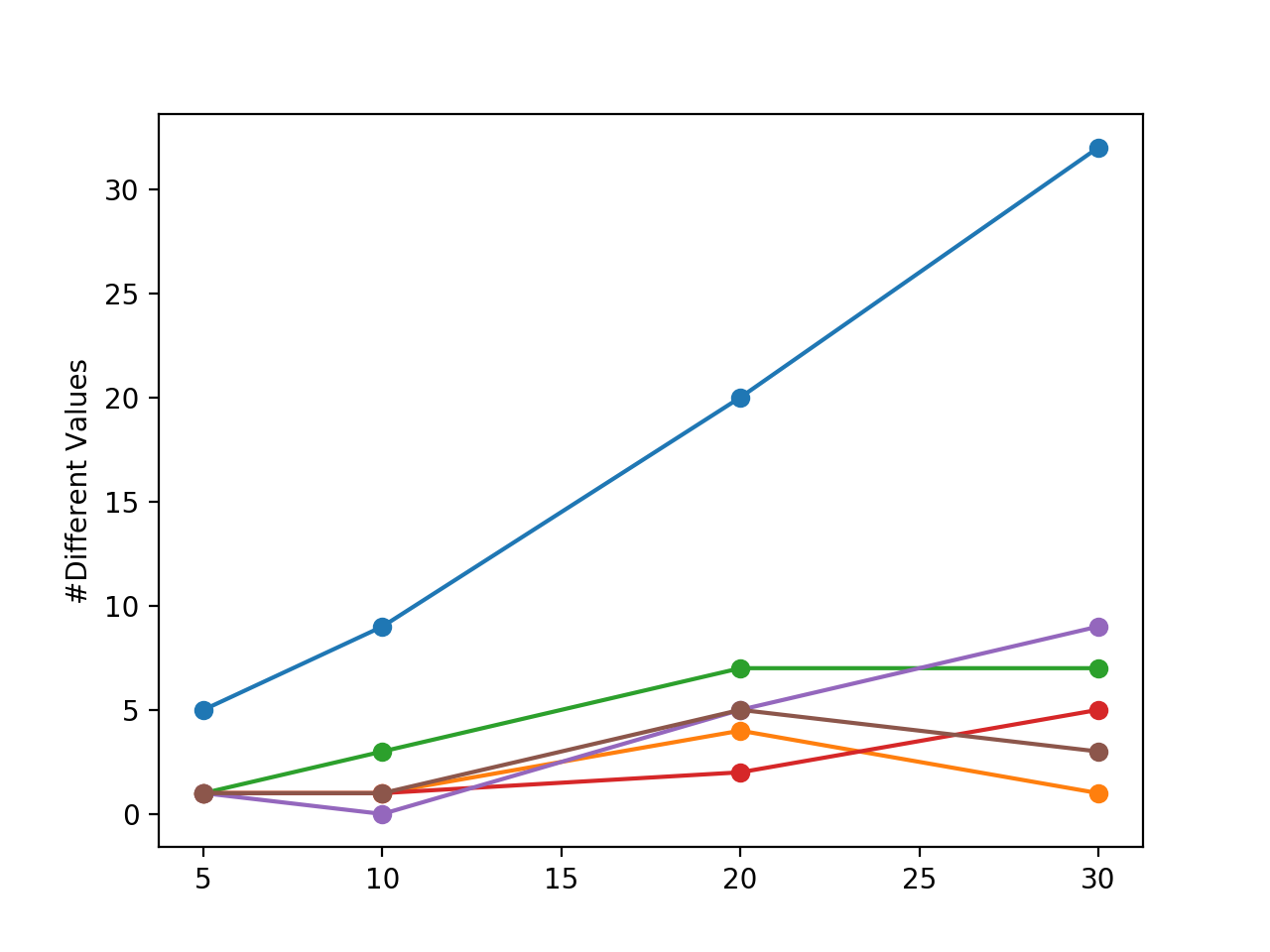}%
    \caption{Iris Dataset}
  \label{fig:sub-third}
\end{subfigure}%
 \begin{subfigure}{.5\linewidth}
  \centering
\includegraphics[width=\textwidth]{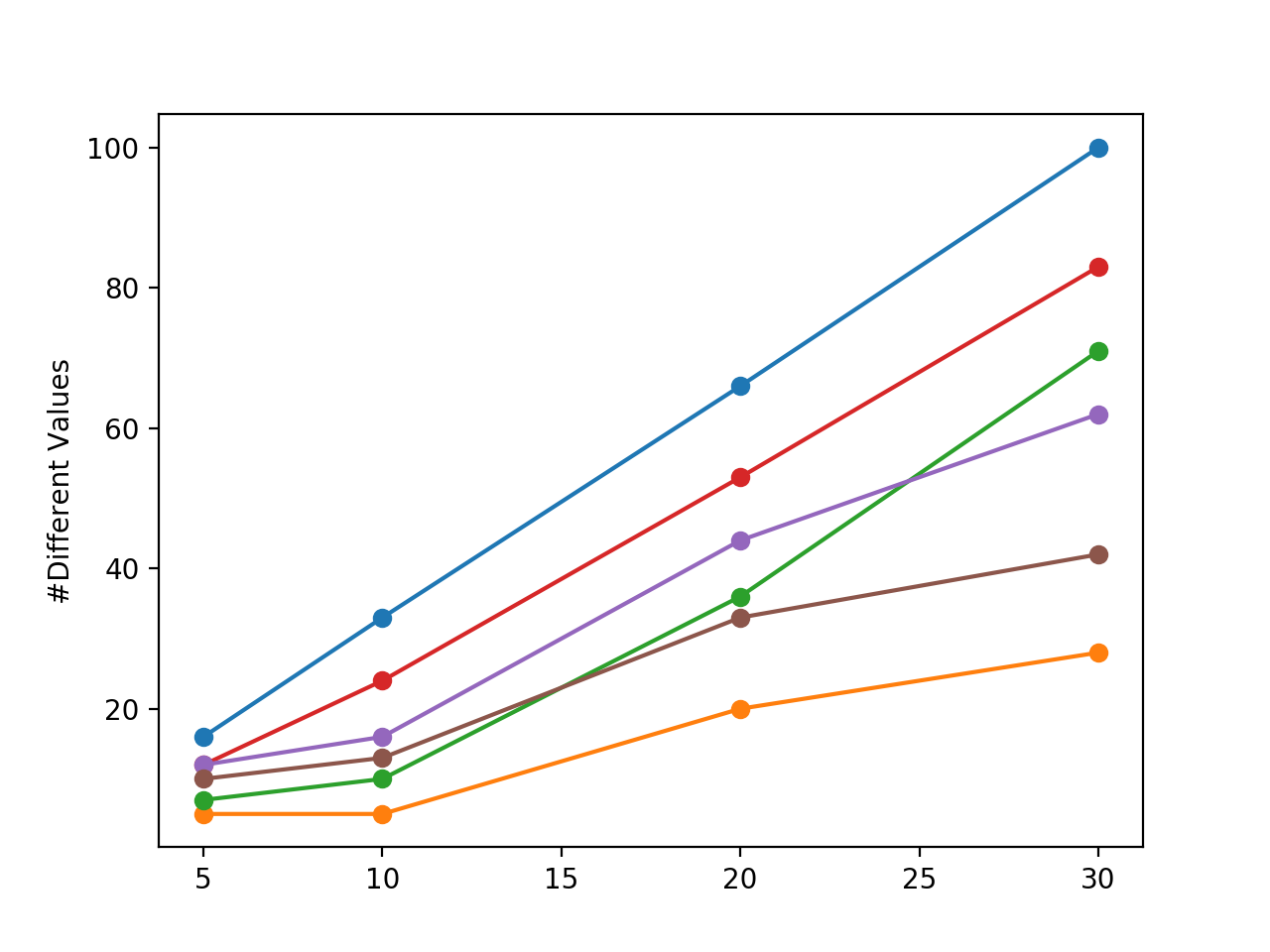}%
  \caption{Ecoli Dataset}
  \label{fig:sub-fourth}
\end{subfigure}
\caption{Number of different values in integer and category columns.}
\label{fig:diff_int_cols}
\end{figure}
\textbf{Number of different values in integer and category columns.} 
The encoded categorical or integer columns require the values only in integer format. However, numerical imputation techniques impute the values for all columns by real numbers. These techniques may give very small root mean square error for these values but are not much useful for the imputation use case. These real values can be converted to integer values by rounding them. 
Figure~\ref{fig:diff_int_cols} shows the number of different values in integer and category columns in four datasets. 


\section{Conclusion}
For the datasets with no or few associations, i.e., attributes are independent, other techniques give better results than imputation using constraints. But when the attributes are related, which is more often than not in most real datasets, our technique gives better results. 

\bibliographystyle{plain}
\bibliography{main}

\end{document}